\documentclass[letterpaper, 10 pt, conference]{ieeeconf}  % Comment this line out if you need a4paper

\IEEEoverridecommandlockouts                              % This command is only needed if 
\usepackage{url}
\usepackage{amsmath} % assumes amsmath package installed
\usepackage{graphicx}
\usepackage{multirow}
\usepackage[pagebackref=true,breaklinks=true,letterpaper=true,colorlinks,bookmarks=false]{hyperref}

\title{\LARGE \bf
Toward Hierarchical Self-Supervised Monocular Absolute Depth Estimation for Autonomous Driving Applications
}

\author{Feng Xue$^{1}$, Guirong Zhuo$^{1,*}$, Ziyuan Huang$^{2}$, Wufei Fu$^{1}$, Zhuoyue Wu$^{1}$ and Marcelo H. Ang Jr$^{2}$% <-this % stops a space
\thanks{$^{1}$Feng Xue, Guirong Zhuo, Wufei Fu and Zhuoyue Wu are with the School of Automotive Studies, Tongji University, 201804 Shanghai, China {\tt\small zhuoguirong@tongji.edu.cn}}%
\thanks{$^{2}$Ziyuan Huang and Marcelo H. Ang Jr are with Department of Mechanical Engineering, National University of Singapore, Singapore
{\tt\small mpeangh@nus.edu.sg}}%
}

\begin{document}

\maketitle
\thispagestyle{empty}
\pagestyle{empty}

%%%%%%%%%%%%%%%%%%%%%%%%%%%%%%%%%%%%%%%%%%%%%%%%%%%%%%%%%%%%%%%%%%%%%%%%%%%%%%%%
\begin{abstract}

In recent years, self-supervised methods for monocular depth estimation has rapidly become an significant branch of depth estimation task, especially for autonomous driving applications. Despite the high overall precision achieved, current methods still suffer from a) imprecise object-level depth inference and b) uncertain scale factor. The former problem would cause texture copy or provide inaccurate object boundary, and the latter would require current methods to have an additional sensor like LiDAR to provide depth ground-truth or stereo camera as additional training inputs, which makes them difficult to implement. In this work, we propose to address these two problems together by introducing DNet. Our contributions are twofold: a) a novel dense connected prediction (DCP) layer is proposed to provide better object-level depth estimation and b) specifically for autonomous driving scenarios, dense geometrical constrains (DGC) is introduced so that precise scale factor can be recovered without additional cost for autonomous vehicles. Extensive experiments have been conducted and, both DCP layer and DGC module are proved to be effectively solving the aforementioned problems respectively. Thanks to DCP layer, object boundary can now be better distinguished in the depth map and the depth is more continues on object level. It is also demonstrated that the performance of using DGC to perform scale recovery is comparable to that using ground-truth information, when the camera height is given and the ground point takes up more than 1.03\% of the pixels. Code is available at \url{https://github.com/TJ-IPLab/DNet}.

\end{abstract}

%%%%%%%%%%%%%%%%%%%%%%%%%%%%%%%%%%%%%%%%%%%%%%%%%%%%%%%%%%%%%%%%%%%%%%%%%%%%%%%%
\section{INTRODUCTION}

Estimating an accurate depth map from single RGB image is of great significance in 3D scene understanding as well as in many real-world applications such as augmented reality and autonomous driving. Compared to traditional hand-crafted feature-based methods~\cite{Karsch2012ECCV}, supervised~\cite{Eigen2014NeurIPS,Fu2018CVPR,Liu2015TPAMI,lee2019big,yin2019enforcing,alhashim2018high} and stereo self-supervised~\cite{goldman2019learn,pillai2019superdepth,poggi2018learning,godard2017unsupervised} learning has been proved to be able to achieve better performance on this task. Unfortunately, these methods either require a large amount of high-quality annotated ground-truth, which is difficult to obtain, or need complex stereo calibration. Therefore, monocular self-supervised learning methods became the focus of research. Some recent works~\cite{Casser2019AAAI,Yang2018CVPR,godard2019digging,Zhou2017CVPR} revealed its great potential to tackle monocular depth estimation task.

\begin{figure}[thpb]
	\centering
	\includegraphics[width=1\columnwidth]{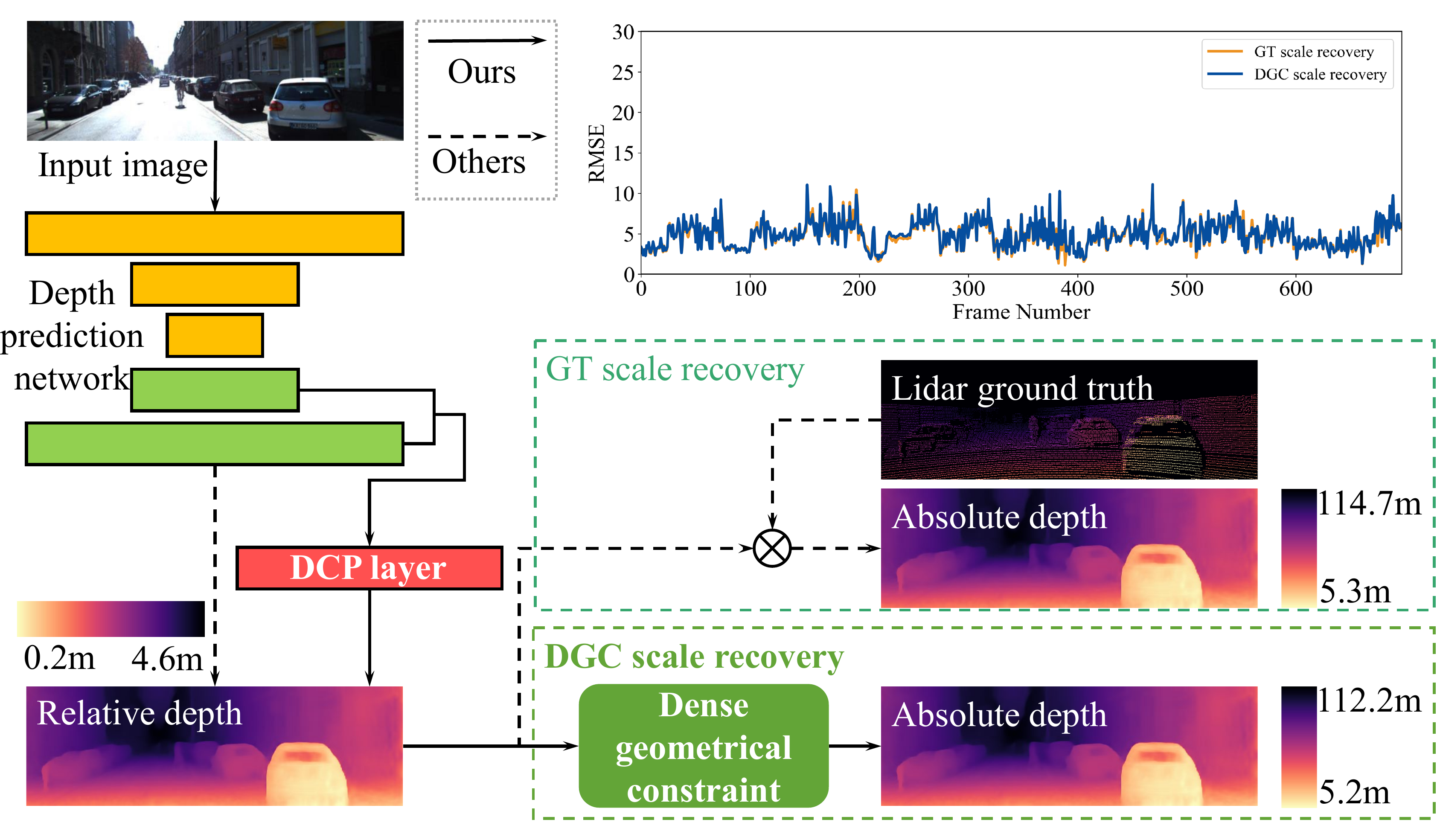}
	\caption{Structure difference of DNet with other self-supervised monocular depth estimation methods. Solid lines indicates our work flow and dotted lines are that of other methods. Dense connected prediction (DCP) layer is introduced to generate hierarchical features for better object-level depth inference, and dense geometrical constraint (DGC) is introduced to directly estimate absolute depth from monocular images. Performance comparison of \textit{DGC} and ground-truth based scale recovery is indicated in the top-right plot.}
	\label{fig1:structure-and-performance-comparison}
\end{figure}

Despite its potential to reach satisfying performances, current methods have two shortcomings. One of them is that they are only able to estimate relative depth rather than the absolute one. For evaluation, scale factor is calculated by ratio between the medians of ground-truth (given by LiDAR) and predicted depth~\cite{Casser2019AAAI,Yang2018CVPR,godard2019digging,Zhou2017CVPR}, as can be seen from Fig.~\ref{fig1:structure-and-performance-comparison}. Theoretically, it is a decent solution. However, for practical uses, obtaining ground-truth in real applications using other sensors not only raises the cost, it also complexes the system, leading to complicated joint calibration processes and synchronization problems.

Another problem is that because the decoder of current methods predict depth in different resolutions separately, some details on object-level is omitted. For example, object boundary can be blurred and the depth of texture on the object may be predicted differently than the object itself.

In this paper, we propose DNet, a novel self-supervised monocular depth estimation pipeline that exploits densely connected hierarchical features to obtain more precise object-level depth inference, and uses dense geometrical constraint to eliminate the dependence on additional sensors or depth ground-truth to perform scale recovery, so that it is easier to be brought into practical use. 

Our contributions are listed as follows:

\begin{itemize}
	\item We improve the former multi-scale estimation strategy by proposing a novel dense connected prediction (DCP) layer. Instead of predicting depth and computing reconstruction loss separately under different scales, the proposed DCP layer exploits hierarchical feature so that object-level depth inference can be made based on multi-scale prediction features, refining object boundary and reducing visual artifacts.
	\item A novel dense geometrical constraints (DGC) module is introduced to perform high-quality scale recovery for autonomous driving. Based on relative depth estimation, DGC module can finish per-pixel ground segmentation and estimate a camera height from every ground point. Statistical method is applied to determine the camera height so that outliers of ground point extraction can be robustly suppressed. Scale factor can be determined through comparision between the given and estimated camera height.
	\item DNet is extensively evaluated on KITTI\cite{Geiger2013IJRR} Eigen Split~\cite{Eigen2014NeurIPS}, where the results not only showed the capability of DCP layer to improve the performance of object-level depth inference, but also proved that DNet using DGC module has competitive performance against those methods using depth ground-truth to determine scale factor. Ablation studies demonstrated module effectiveness as well as sensitivity of DNet to ground points ratio.
\end{itemize}

\section{RELATED WORKS}

\subsection{Self-supervised monocular depth estimation}

Monocular depth estimation has always been an important aspect of scene understanding. Some works apply supervised~\cite{Eigen2014NeurIPS,Fu2018CVPR,Liu2015TPAMI,lee2019big,yin2019enforcing,alhashim2018high} or stereo self-supervised~\cite{goldman2019learn,pillai2019superdepth,poggi2018learning,godard2017unsupervised} methods to tackle the problem. However, due to the difficulty of obtaining large amount of labeled data or complex stereo calibration to train the depth estimation network, monocular self-supervised method was proposed instead~\cite{Casser2019AAAI,Yang2018CVPR,godard2019digging,Zhou2017CVPR}. 

Proposed by the pioneering work~\cite{Zhou2017CVPR}, the basic idea is to use photometric reconstruction loss calculated by comparing the target image with the target view reconstructed from nearby source views. However, it assumes that the scene is static and that no occlusion is present between different consecutive frames. \cite{Luo2018arXiv,Yin2018CVPR,Vijayanarasimhan2017arXiv} explicitly established different motion models to resolve the moving scene problem. \cite{Yang2017arXiv} introduced 3D surface normal by constructing two additional layers for better depth estimation. \cite{godard2019digging} replaced the original photometric reconstruction error with per-pixel minimum reprojection error, which partially enabled it to tackle occlusion. It also used up-sampling and proposed auto-masking of stationary pixels to avoid 'holes' of infinite depth generated by low-texture and moving objects respectively. 

However, all aforementioned works predict only relative depth, which means there still exists a scale gap between the prediction and true depth. For evaluation purpose, ratio between medians of ground-truth and current prediction is employed to acquire absolute depth. Unfortunately, in real application scenarios, ground-truth is either too difficult or financially expensive to obtain. Therefore, a scale recovering approach which is free of depth ground truth is called for.

\subsection{Monocular scale recovery}
% To read
Scale uncertainty has always been a problem for 3D vision for monocular camera. To recover scale factor and achieve absolute depth estimation, \cite{pinard2018learning} utilizes pose information and \cite{roussel2019monocular} uses stereo data to pretrain network, both introducing additional sensor information but the results were no as satisfactory. Besides depth estimation, a typical example of this is monocular visual SLAM. In order to mitigate this, \cite{frost2016object,sucar2017probabilistic} integrate object detection algorithms into monocular visual SLAM system and take advantage of object size prior to recover scale. However, in addition to the significant increase of computation complexity, these methods show limited robustness under scenes without known object classes.

\begin{figure*}[thpb]
	\centering
	\includegraphics[width=1\linewidth]{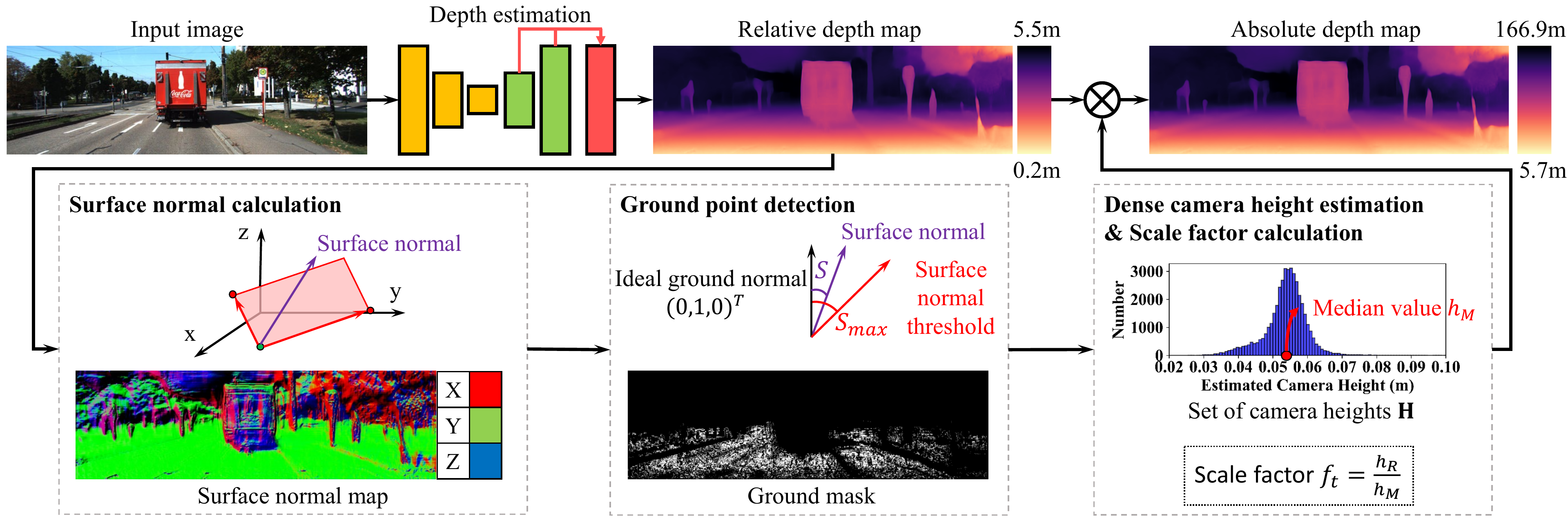}
	\caption{Overall structure of proposed DNet pipeline. The algorithm first estimates relative depth using the depth estimation network with our proposed DCP layer (pink layer in the figure) to generate and exploit hierarchical features. After that, scale recovery module pops in. It utilizes geometrical relationship between the ground and the camera, using extracted ground points to densely calculate camera heights point by point. Median value of all camera heights is then selected to be the final estimated value and used to obtain scale factor. Combined with relative depth map generated in the first step, scale recovery module outputs absolute depth of the given monocular image.}
	\label{fig2:overall-structure}
\end{figure*}

Handling the geometrical relationship between camera and ground is also an effective approach to tackle this problem. This geometrical constrain is broadly used in autonomous driving tasks, for ground is commonly seen in images captured by on-board cameras. The main task of these methods is to estimate a relative camera height using camera-ground geometrical constrains, and thus infer scale with absolute camera height prior. \cite{Choi2011ICCAR} extracted ground using trained classifier, but it doesn't possess an excellent generalization power. \cite{Song2015TPAMI} extracted the ground points densely in region of interest similar to \cite{kitt2011monocular,zhou2019ground}, but it requires dense stereo to be added to the system, which can potentially raise cost and increase complexity. In \cite{Wang2018ICRA}, the most similar work to this one, used surface normal to extract ground points and thus calculate camera height. However, due to the sparsity originated in its key-point-based strategy, data association through consecutive frames are needed, makes this method hardly integrated into monocular depth estimation tasks which use only single image as input. In additon, this method regards the ground as a whole, flat panel with single surface normal, which is a strong assumption for autonomous driving scenarios. In contrast, our method is free of data association, which means it can be integrated into both monocular depth estimation and visual SLAM tasks. Furthermore, our method achieve per-pixel surface normal calculation and ground segmentation, makes the algorithm robust to different road conditions for autonomous driving.

\section{METHOD}

In this section, a novel pipeline called DNet specifically designed for monocular absolute depth estimation in autonomous driving applications is proposed. The pipeline can be divided into two parts, respectively relative depth estimation, with dense connected prediction (DCP) layer to improve object-level depth inference, and scale recovery based on dense geometrical constraint, without needing any additional sensor signals or depth ground-truth. The overview of DNet can be seen in Fig.~\ref{fig2:overall-structure}.

\subsection{Relative depth estimation} 

The proposed DNet is based on Monodepth2~\cite{godard2019digging}. As all self-supervised depth estimation methods, its object-level inference can still have texture copy and imprecise object boundaries. In this section, we will first introduce Monodepth2 and then resolve this issue by introducing DCP layer to replace full resolution module used in Monodepth2.

\subsubsection{Baseline: Monodepth2 w/o full resolution}

\textbf{Architecture: }Two networks are required in monocular self-supervision architecture, respectively a depth network and a pose network. Single image $I_t$ of the $t$-th frame is taken as the input of the depth network. Depth network outputs a dense relative depth map $\mathbf{D}^{rel}_t$. Pose network takes $\lbrace \mathbf{I}_{t-1}, \mathbf{I}_t\rbrace $ and $\lbrace \mathbf{I}_t, \mathbf{I}_{t+1}\rbrace $ sequentially as inputs and then outputs camera poses of the $t$-th image relative to that of the $(t-1)$-th and $(t+1)$-th images, i.e., $\lbrace \mathbf{T}^{rel}_{t\rightarrow t-1}, \mathbf{T}^{rel}_{t\rightarrow t+1} \rbrace$. 

\textbf{Self-supervision loss: }Two parts constitute the overall loss, respectively per-pixel minimum reconstruction loss $L_p$ and inverse depth smoothness loss $L_s$. Reconstruction loss is calculated by firstly inverse warping source images $\lbrace \mathbf{I}_{t-1}, \mathbf{I}_{t+1}\rbrace $ to rebuild two target images $\lbrace \mathbf{I}_{t-1\rightarrow t}, \mathbf{I}_{t+1\rightarrow t} \rbrace$. After that, photometric error (PE) between reconstructed image and target image is calculated combining structural similarity index (SSIM)~\cite{Wang2004TIP} and L1 norm between two images $\mathbf{I}_a, \mathbf{I}_b$ as follows:
\begin{equation}\label{pe}
\text{PE}(\mathbf{I}_a, \mathbf{I}_b) = \alpha \frac{1 - \text{SSIM}(\mathbf{I}_a, \mathbf{I}_b)}{2} + (1-\alpha) \Vert \mathbf{I}_a - \mathbf{I}_b\Vert_1 \ ,
\end{equation}
\noindent where $\alpha$ is used for weight adjustment. 

Per-pixel minimum loss $\mathbf{L}_p$ is then calculated as follows:
%to solve occlusion situation and 'holes' of infinite depth:
\begin{equation}\label{lp}
\mathbf{L}_p = \min_{\mathbf{I}'}\left( PE(\mathbf{I}', \mathbf{I}_t)\right) \ ,
\end{equation}
\noindent where $\mathbf{I}'\in \lbrace \mathbf{I}_{t-1\rightarrow t}, \mathbf{I}_{t+1\rightarrow t}, \mathbf{I}_{t-1}, \mathbf{I}_{t+1} \rbrace$.

Combined with edge-aware smoothness loss $\mathbf{L}_s$:
\begin{equation}\label{smoothness}
\mathbf{L}_s = |\partial_x d_t^*|e^{-|\partial_x\mathbf{I}_t|} + |\partial_y d_t^*|e^{-|\partial_y\mathbf{I}_t|} \ ,
\end{equation}
\noindent where $d^{*}_t = d_t/\bar{d_t}$ is the mean-normalized inverse depth, overall loss can be constructed with two hyper-parameters $\mu$ and $\lambda$ as:
\begin{equation}\label{overall_loss_function}
\mathbf{L}_{i} = \sum_{i} (\mu \mathbf{L}_{p,i} + \lambda w_i \mathbf{L}_{s,i}) \ , 
\end{equation}
\noindent where subscript $i$ denotes different resolution layers of the decoder. $w_i$ is determined according to the resolution. 

\subsubsection{DNet \& Dense connected prediction layer}

\textbf{Overall loss: }Because the photometric error of low resolution depth prediction can be the result of wrong network prediction or the aliasing of down-sampling, using the same weight in loss for low-res and high-res results can mislead the network to converge in non-optimal values. Additionally, in consideration that features with lower resolution are reused for multiple times, the weight of error in lower resolution depth prediction is reduced as follows: 
\begin{equation}\label{overall_loss_function_ours}
\mathbf{L}_{i} = \sum_{i} (\mu v_i \mathbf{L}_{p,i} + \lambda w_i \mathbf{L}_{s,i}) \ . 
\end{equation}
\noindent where $v_i < 1$ is introduced as weight adjustment parameter.

\textbf{DCP layer: }In order to handle local gradient caused by bilinear sampling~\cite{jaderberg2015spatial} and local minima, current works~\cite{Casser2019AAAI,Yang2018CVPR,godard2019digging,Zhou2017CVPR} including our baseline Monodepth2 use multi-scale depth prediction strategy. This strategy implicitly uses low-res features to predict depth by repeated upsampling layers, which has the tendency of depth artifacts (Fig.~\ref{fig:DCP-layer-qualitative-results}). Motivated by reducing the depth artifacts and acquiring more reasonable object-level depth inference, we propose a novel DCP layer that explicitly combines features in different scales hierarchically. The intuition is based on the observation that low-res layers of decoder network can provide more reliable object-level depth inference and high-res layers focus more on local depth details. 

\begin{figure}[thpb]
	\centering
	\includegraphics[width=1\columnwidth]{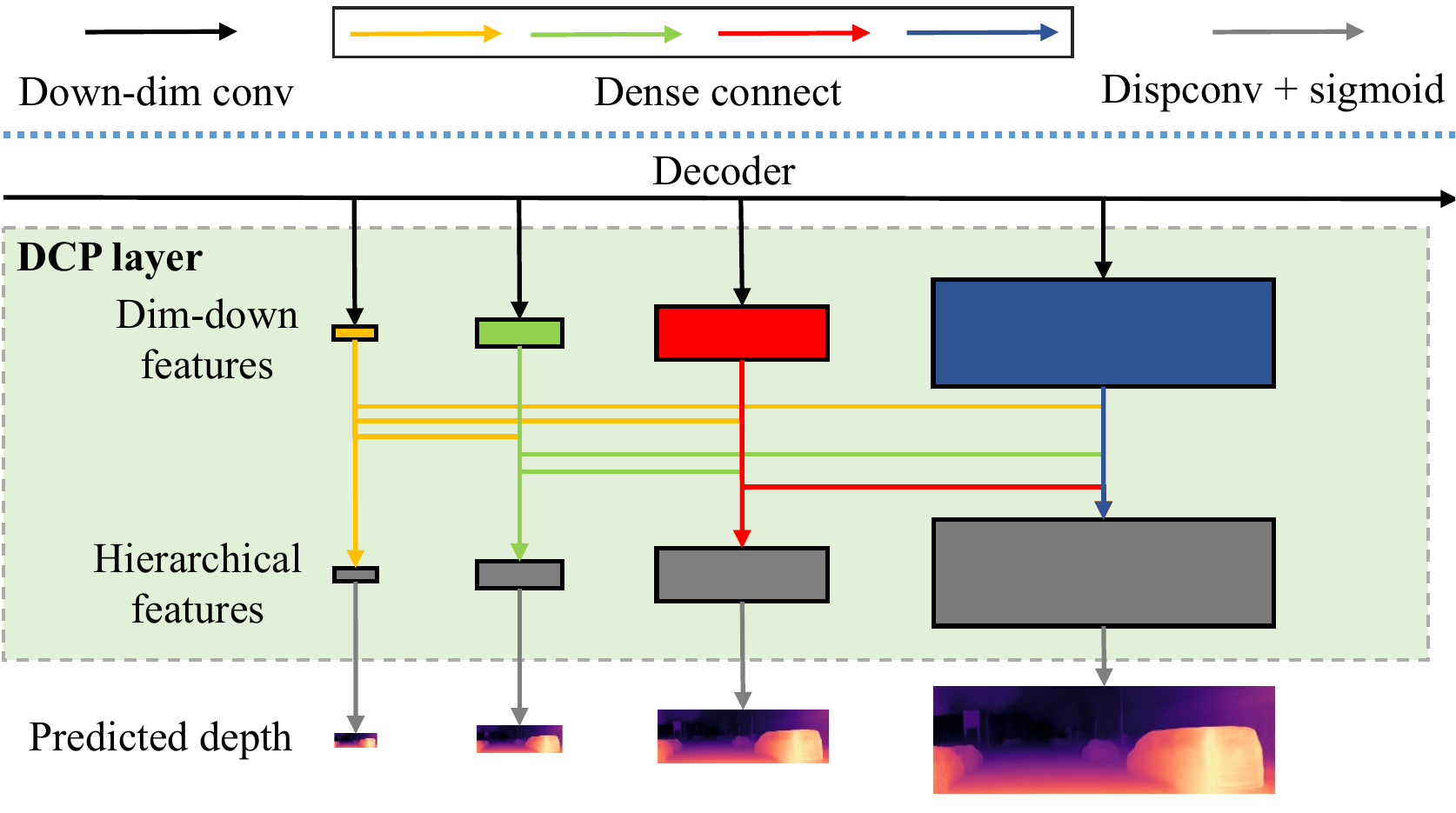}
	\caption{Structure of proposed DCP layer. Different from baseline multi-scale prediction strategy (directly uses feature in different resolutions independently), features are densely connected from low to high resolution to form hierarchical features in DCP layer. }
	\label{fig3:Difference between DCP layer and baseline}
\end{figure}

Formally, the numbers of feature channels in different scales are reduced to eight using a convolutional layer in the DCP layer, so that the number of channels are uniformed and calculations afterwards can be simplified. Features in low-res layers are then up-sampled and concatenated to higher-res layer features. By doing this, we introduce more precise object-level inference into higher resolution depth predictions that originally care less about object-level depth. The final depth estimation is performed based on the hierarchical features provided by densely connected feature layers. Detailed structure can be seen in Fig.~\ref{fig3:Difference between DCP layer and baseline}. 

\subsection{Scale recovery}

Scale recovery is performed after relative depth is predicted so that absolute depth map can be generated solely relying on monocular image. Dense geometrical constraint (DGC) is thus introduced. DGC is specifically designed for autonomous driving applications. It works under the assumption that there are enough ground points in the monocular image, which is usually the case for autonomous driving. Unlike the scale recovery employed by feature-based visual odometry, ground points are densely extracted by DGC from the monocular images to form a dense ground point map. Each point in the map is used to estimate one camera height, as can be seen in Fig.~\ref{fig:road-model}. A large number of camera heights can thus be obtained. By applying statistical methods for overall camera height estimation, outliers can barely harm the estimation result of the scale factor.  

\subsubsection{Surface normal calculation}

The first step is to determine a surface normal for each pixel in the input image. All the pixel points need to be projected to 3D space according to the following equation:
\begin{equation}
\mathbf{D}^{rel}_t(\mathbf{p}_{i,j}) \mathbf{p}_{i,j} = \mathbf{K} \mathbf{P}_{i,j}\ , 
\end{equation}
\noindent where $\mathbf{p}_{i,j}=[i,j,1]^{\top}$ refers to the pixel on the $i$-th row and the $j$-th column in 2D space with one homogeneous coordinate, and $\mathbf{P}_{i,j}=[X,Y,Z]^{\top}$ is the corresponding 3D point, $\mathbf{D}^{rel}_t(\mathbf{p}_{i,j})$ is the depth of that specific point, and $\mathbf{K}$ is the camera intrinsic matrix. 

\begin{figure}[thpb]
	\centering
	\includegraphics[width=1\linewidth]{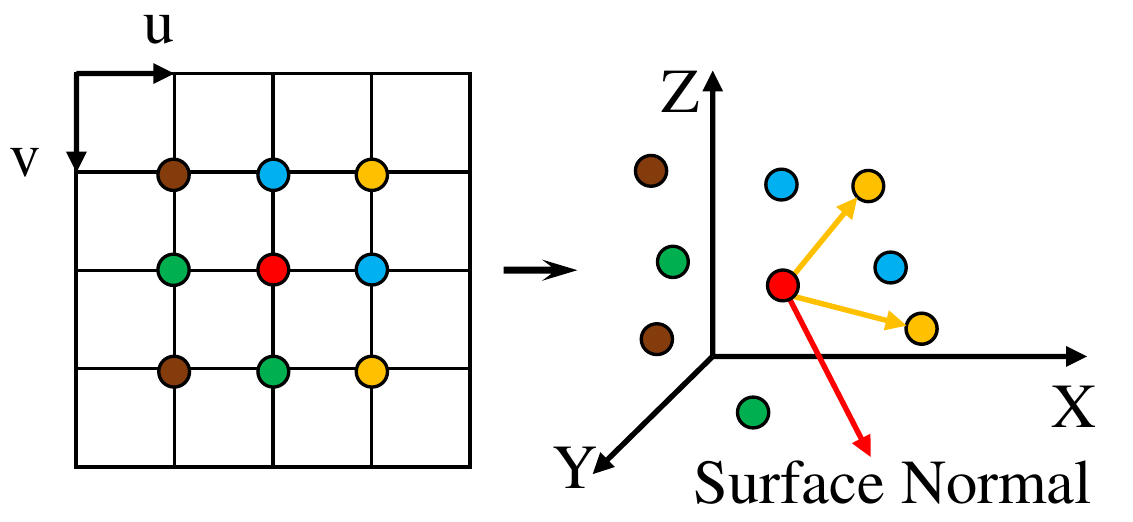}
	\caption{2D to 3D projection and pairing of 8-neighbors in surface normal calculation. Points with the same color is paired to form two vectors respectively with the center point. Four surface normals can be calculated from four vector pairs and used to form one surface normal.}
	\label{fig:surface-normal-calculation}
\end{figure}

Similar to \cite{Yang2017arXiv}, for each pixel point , 8-neighbor convention is used to determine several planes around it, as in Fig.~\ref{fig:surface-normal-calculation}. All 8 neighbors of $\mathbf{p}_{i,j}$ are grouped into 4 pairs. Two vectors of $\mathbf{p}_{i,j}$ connected respectively to two points in one pair form a 90-degree angle, i.e., $G(\mathbf{P}_{i,j}) = \lbrace \lbrack\mathbf{P}_{i+1,j}, \mathbf{P}_{i,j-1}\rbrack, \lbrack \mathbf{P}_{i+1,j-1}, \mathbf{P}_{i-1,j-1} \rbrack ...\rbrace$. Four pairs of vector constitutes 4 surfaces, thus generating 4 surface normals, which can be calculated by:
\begin{equation}
\mathbf{n}_g = \overrightarrow{\mathbf{P}_{i,j}G_{g,1}} \times \overrightarrow{\mathbf{P}_{i,j}G_{g,2}}\ ,
\end{equation}
\noindent where $G_{a,b}$ denotes the $b$-th element of the $a$-th pair in $G(\mathbf{P}_{i,j})$ and $g = 1,2,3,4$. 

The final normalized surface normal of point $\mathbf{P}_{i,j}$ is given by normalizing and averaging four estimated normals:
\begin{equation}
\mathbf{N}(\mathbf{P}_{i,j}) = \frac{\sum_{g} \mathbf{n} / \Vert \mathbf{n}_g\Vert_2}{4}\ .
\end{equation}

\subsubsection{Ground point detection}

Ground points usually refers to the points that has a normalized normal close to ideal ground normal, i.e., $\mathbf{\tilde{n}} = (0,1,0)^{\top}$. With this ideal target normal and the calculated normalized surface normal, we propose a similarity function $s(\mathbf{P}_{i,j})$ based on absolute value of cosine function. The calculated similarity $S$ can be used as a simple criteria to determine whether $\mathbf{P}_{i,j}$ is a ground point or not.

\begin{figure}[thpb]
	\centering
	\includegraphics[width=1\linewidth]{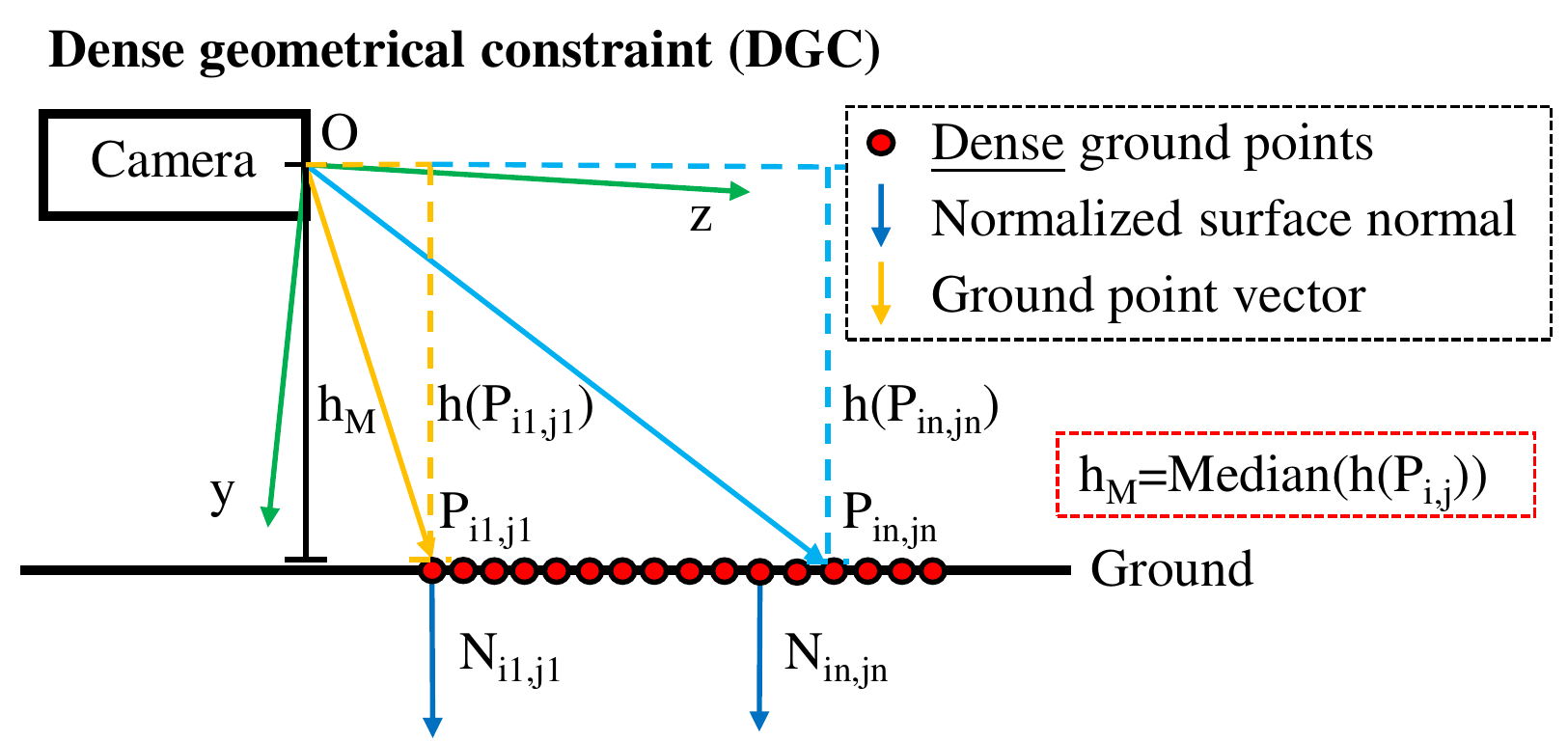}
	\caption{Schematic for DGC. Different from common geometrical constraint, which outputs only one surface normal for all ground points, for each ground point in DGC, a surface normal vector is calculated. Each surface normal vector is used to estimate one camera height. The overall camera height is estimated through calculating the median of all estimated camera heights.}
	\label{fig:road-model}
\end{figure}

\begin{table*}[h]
	\caption{\textbf{Quantitative results.} Performance of DNet pipeline compared to existing methods. The experiment results of existing methods are from respective papers. For scale factor, GT is using LIDAR depth ground truth, P is using additional pose information, S is using stereo pretrained network and DGC is our proposed dense geometrical constrain method. \textbf{Bold} and \underline{underlined} data denotes the best and second best performance respectively.}
	\label{tab:absolute_evaluation}
	\begin{center}
		\begin{tabular}{|l|c|c|c|c|c|c|c|c|}
			\hline
			\multirow{2}*{Method}& Scale & \multicolumn{4}{c|}{Lower is better} & \multicolumn{3}{c|}{Higher is better}\\
			\cline{3-9}
			~ & Factor & Abs Rel & Sq Rel & RMSE & RMSE log & $\delta<1.25$ & $\delta<1.25^2$ & $\delta<1.25^3$ \\
			\hline
			Zhou et al.~\cite{Zhou2017CVPR}CVPR'17 & GT & 0.183 & 1.595 & 6.709 & 0.270 & 0.734 & 0.902 & 0.959 \\
			Yang et al.~\cite{Yang2017arXiv}AAAI'18 & GT & 0.182 & 1.481 & 6.501 & 0.267 & 0.725 & 0.906 & 0.963 \\
			Mahjourian et al.~\cite{Mahjourian2018CVPR}CVPR'18 & GT & 0.163 & 1.240 & 6.220 & 0.250 & 0.762 & 0.916 & 0.968 \\
			LEGO~\cite{Yang2018CVPR}CVPR'18 & GT & 0.162 & 1.352 & 6.276 & 0.252 & - & - & - \\
			DDVO~\cite{Wang2018CVPR}CVPR'18 & GT & 0.151 & 1.257 & 5.583 & 0.228 & 0.810 & 0.936 & 0.974 \\
			DF-Net~\cite{Zou2018ECCV}ECCV'18 & GT & 0.150 & 1.124 & 5.507 & 0.223 & 0.806 & 0.933 & 0.973 \\
			GeoNet~\cite{Yin2018CVPR}CVPR'18 & GT & 0.149 & 1.060 & 5.567 & 0.226 & 0.796 & 0.935 & 0.975 \\
			EPC++~\cite{Luo2018arXiv}TPAMI'18 & GT & 0.141 & 1.029 & 5.350 & 0.216 & 0.816 & 0.941 & 0.976 \\
			Struct2Depth~\cite{Casser2019AAAI}AAAI'19 & GT & 0.141 & 1.026 & 5.291 & 0.215 & 0.816 & 0.945 & \underline{0.979} \\
			CC~\cite{Ranjan2019CVPR}CVPR'19 & GT & 0.139 & 1.032 & 5.199 & 0.213 & 0.827 & 0.943 & 0.977 \\
			Bian et al.~\cite{bian2019unsupervised}NIPS'19 & GT & 0.128 & 1.047 & 5.234 & 0.208 & \underline{0.846} & 0.947 & 0.976 \\
			Monodepth2~\cite{godard2019digging}ICCV'19 & GT & \underline{0.115} & \underline{0.903} & \underline{4.863} & \underline{0.193} & \textbf{0.877} & \underline{0.959} & \textbf{0.981} \\
			\textbf{DNet (Ours)} & GT & \textbf{0.113} & \textbf{0.864} & \textbf{4.812} & \textbf{0.191} & \textbf{0.877} & \textbf{0.960} & \textbf{0.981} \\ 
			\hline
			Pinard et al.~\cite{pinard2018learning}ECCV'18 & P & 0.271 & 4.495 & 7.312 & 0.345 & 0.678 & 0.856 & 0.924 \\
			Roussel et al.~\cite{roussel2019monocular}IROS'19 & S & \underline{0.175} & \underline{1.585} & \underline{6.901} & \underline{0.281} & \underline{0.751} & \underline{0.905} & \underline{0.959} \\
			\textbf{DNet (Ours)} & DGC & \textbf{0.118} & \textbf{0.925} & \textbf{4.918} & \textbf{0.199} & \textbf{0.862} & \textbf{0.953} & \textbf{0.979} \\
			\hline
		\end{tabular}
	\end{center}
\end{table*}

\begin{equation}
S = s(\mathbf{P}_{i,j}) = |\angle(\mathbf{\tilde{n}},\mathbf{N}(\mathbf{P}_{i,j}))| = |arccos\frac{\mathbf{\tilde{n}}\cdot\mathbf{P}_{i,j}} {\Vert\mathbf{\tilde{n}}\Vert \Vert\mathbf{P}_{i,j}\Vert}| \ ,
\end{equation}
\noindent where operator $\cdot$ denotes the inner product operation. 

Considering the uncertainty produced by estimating the surface normal and the y-axis of camera coordinate system is not strictly perpendicular to the ground as in Fig.~\ref{fig:road-model}, a threshold $S_{max}$ is set. For $S<S_{max}$, the pixel point is considered as ground points. After determination for ground points has finished for all pixel points, a set of ground points $\mathbf{GP} = \lbrace\mathbf{P}_{i,j} | s(\mathbf{P}_{i,j})<S_{max}, y(\mathbf{P}_{i,j}) > 0\rbrace$ is detected, where $y(\mathbf{P}_{i,j})$ denotes the y-axis value of $\mathbf{P}_{i,j}$. A ground mask is thereafter generated.

\subsubsection{Camera height estimation}

When all the ground points have been densely identified from the image, the geometrical relationship between ground points and camera itself is ready to be exploited. As can be seen from Fig.~\ref{fig:road-model}, camera height is the projection of vector $\overrightarrow{\mathbf{OP}_{i,j}}$ in the direction of surface normal of point $\mathbf{P}_{i,j}$, i.e., $\mathbf{N}(\mathbf{P}_{i,j})$. Therefore, camera height of $\mathbf{P}_{i,j}$ can be calculated as follows:
\begin{equation}
h(\mathbf{P}_{i,j}) = \mathbf{N}(\mathbf{P}_{i,j})^{\top} \cdot \overrightarrow{\mathbf{OP}_{i,j}} \ ,
\end{equation}
\noindent where $\overrightarrow{\mathbf{OP}_{i,j}} = \mathbf{P}_{i,j} = [X, Y, Z]^{\top}$. This operation is done for all $ \mathbf{P}_{i,j} \in \mathbf{GP}$.

Now a set of camera heights $\mathbf{H} = \lbrace h(\mathbf{P}_{i,j}) | \mathbf{P}_{i,j} \in \mathbf{GP}\rbrace$ with element number equal to that of ground points is obtained. But for overall scale factor, one single camera height should be estimated for the relative depth map. After careful experiments, median of all estimated camera heights $h_M = Median(\mathbf{H})$ is selected as the final camera height. 

\subsubsection{Scale factor calculation}

Given the camera height estimated for current relative depth map for $\mathbf{I}_t$, in order to calculate the scale factor, all that is still needed is the real height of the camera $h_R$. The scale factor for the current relative depth estimation is simply determined as follows:
\begin{equation}
f_t = \frac{h_R}{h_M} \ .
\end{equation}

\subsection{Absolute depth estimation}

After successfully estimated the scale factor for current relative depth map $\mathbf{D}^{rel}_t$, absolute depth can be thus pixel-wise calculated:
\begin{equation}
\mathbf{D}^{abs}_t = f_t \mathbf{D}^{rel}_t.
\end{equation}
\noindent where $\mathbf{D}^{abs}_t$ denotes the absolute depth estimated for current image $\mathbf{I}_t$.

\section{EXPERIMENT}

Thorough experiments are presented here for evaluation of DNet pipeline. Quantitative results show our proposed DNet is able to achieve competitive performance on both relative depth estimation and scale recovery. Also, ablation study is performed to prove the effectiveness of our proposed DCP layer. And due to the dependency of enough visible ground,  experiments under different ground point ratio show the robustness of DGC scale recovery module.

\subsection{Implementation details}

The same training parameters and method as Monodepth2 are used. Specifically, we set $\mu = 1, \lambda = 0.001$, and $\alpha$ for SSIM is equal to $0.85$. Only monocular image sequence is used during training. For scale recovery, angle threshold $S_{max} = 5$. Low values are assigned to $v_i$ and $w_i$ for low-res predictions, i.e., $\mathbf{v} = \mathbf{w} = \{1/8, 1/4, 1/2, 1\}$.

The experiments are run on a computer with Intel Xeon 8163 CPU (2.5GHz) and NVIDIA RTX 2080 Ti. 

\subsection{Evaluation dataset}

All experiments for evaluation of DNet are conducted on the Eigen split~\cite{Eigen2014NeurIPS} of KITTI\cite{Geiger2013IJRR} 2015 containing 697 test images. For evaluation of depth estimation results, it contains ground truth projected from LiDAR 3D point clouds to 2D depth maps. However, there is no ground truth for scale factors to transfer relative depth maps to absolute depth maps. Usually used method is to use the ratio between medians of LiDAR detected depth values and estimated ones as ground truth of scale factor. 

\subsection{Quantitative evaluation}

Thorough quantitative evaluation is presented to show the overall performance of DNet pipeline on both relative and absolute depth estimation. Commonly used metrics are adopted for evaluation.

Table~\ref{tab:absolute_evaluation} demonstrates the overall depth estimation performance of DNet, both using ground-truth (GT) and DGC scale recovery, in comparison with 14 self-supervised monocular depth estimators. DNet with GT scale recovery is first evaluated to demonstrate its relative depth estimation performance. As can be seen from the table, DNet with GT scale recovery has achieved a satisfactory result. It has improved compared to Monodepth2 on former four metrics by respectively 1.74\%, 4.32\%, 1.05\% and 1.04\%. 

In terms of absolute depth estimation, DGC performs almost as well as GT scale recovery. Compared to Roussel et al.\cite{roussel2019monocular}, DNet achieves improvement on former four metrics by respectively 32.57\%, 41.64\%, 28.73\% and 29.18\%. The performance of DGC module can even outperform most early depth estimator using GT scale recovery. These indicate that DGC scale recovery method, in spite of its simplicity, can carry out a satisfactory scale recovery. 

\subsection{Ablation study}

In order to better show the benefit of our proposed modules and the robustness against ground point ratio, comprehensive ablation study is conducted. 

\begin{table}[h]
    \scriptsize
	\setlength{\tabcolsep}{0.6mm}
	\caption{\textbf{Ablation study.} Comparison on the performance between baseline and our DNet with proposed DCP layer. Scale factor is determined using LIDAR ground truth.}
	\label{tab:Ablation study}
	\begin{center}
		\begin{tabular}{|l|c|c|c|c|c|c|c|}
			\hline
			\multirow{2}*{Method} & 
			\multicolumn{4}{c|}{Lower is better} & \multicolumn{3}{c|}{Higher is better}\\
			\cline{2-8}
			~ & Abs Rel & Sq Rel & RMSE & RMSE log & $\delta<1.25$ &  $\delta<1.25^2$ & $\delta<1.25^3$ \\
			\hline
			Baseline & 0.117 & 0.894 & 4.899 & 0.195 & 0.871 & 0.958 & \textbf{0.981}
			\\
			\textbf{Ours} & \textbf{0.113} & \textbf{0.864} & \textbf{4.812} & \textbf{0.191} & \textbf{0.877} & \textbf{0.960} & \textbf{0.981} \\
			\hline
		\end{tabular}
	\end{center}
\end{table}

\begin{table}[h]
    \scriptsize
	\setlength{\tabcolsep}{0.6mm}
	\caption{\textbf{Ablation study.} Comparison on the object level prediction performance between baseline and our DNet with proposed DCP layer. Scale factor is determined using LIDAR ground truth.}
	\label{tab:Quantitative results on DCP layer}
	\begin{center}
		\begin{tabular}{|l|c|c|c|c|c|c|c|}
			\hline
			\multirow{2}*{Method} & 
			\multicolumn{4}{c|}{Lower is better} & \multicolumn{3}{c|}{Higher is better}\\
			\cline{2-8}
			~ & Abs Rel & Sq Rel & RMSE & RMSE log & $\delta<1.25$ &  $\delta<1.25^2$ & $\delta<1.25^3$ \\
			\hline
			Baseline & 0.227 & 3.680 & 8.430 & 0.327 & 0.690 & 0.857 & 0.924
			\\
			\textbf{Ours} & \textbf{0.202} & \textbf{2.817} & \textbf{7.941} & \textbf{0.310} & \textbf{0.725} & \textbf{0.875} & \textbf{0.932} \\
			\hline
		\end{tabular}
	\end{center}
\end{table}

\subsubsection{Benefit of DCP layer}In order to show the effectiveness of hierarchical feature generated by DCP layer, comparisons are made between baseline and DNet as can be seen in Table~\ref{tab:Ablation study}. It can be seen that,  our proposed DCP layer can boost the performance on the former four metrics by respectively 3.42\%, 3.36\%, 1.78\%, 2.05\%.

\begin{figure}[thpb]
	\centering
	\begin{minipage}{0.23\textwidth}
		\centering
		\includegraphics[width=1\columnwidth]{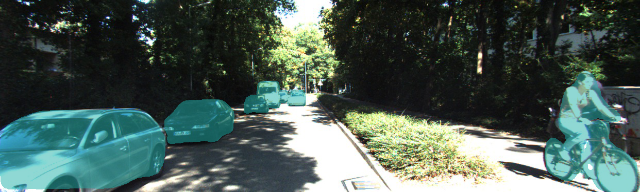}
		\\
		\includegraphics[width=1\columnwidth]{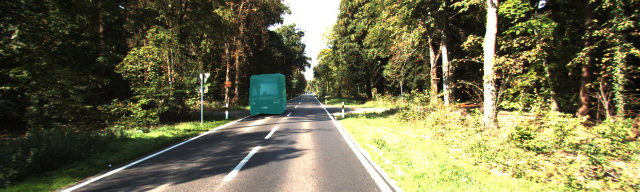}
	\end{minipage}
	\begin{minipage}{0.23\textwidth}
		\centering
		\includegraphics[width=1\columnwidth]{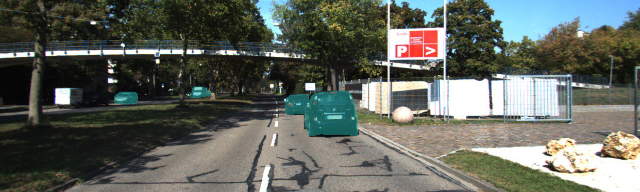}
		\\
		\includegraphics[width=1\columnwidth]{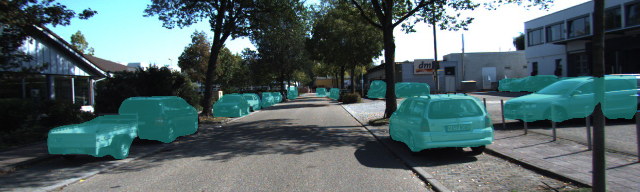}
	\end{minipage}	
	\caption{\textbf{Object masks} extracted by Mask-RCNN\cite{he2017mask}, object level performance is calculated only on pixels in the mask.}
	\label{fig:object masks}
\end{figure}

\subsubsection{Benefit of DCP layer on object-level prediction}Depth estimation on objects can be challenging for the irrgular boundary and texture copy effects. To show the improvement of DCP layer on object-level prediction, Mask-RCNN\cite{he2017mask} is used to generate object masks as shown in Fig.\ref{fig:object masks} on test files and error metrics are calculated only within the masked areas. Table~\ref{tab:Quantitative results on DCP layer} compares performance between baseline and DNet on the object-level depth prediction. Our proposed DCP layer improves the object-level prediction performance on the former four metrics by respectively 11.01\%, 23.45\%, 5.80\%, 2.14\%.

\subsection{Robustness of DGC scale recovery against visible ground: }
\begin{figure}[thpb]
	\centering
	\includegraphics[width=1\columnwidth]{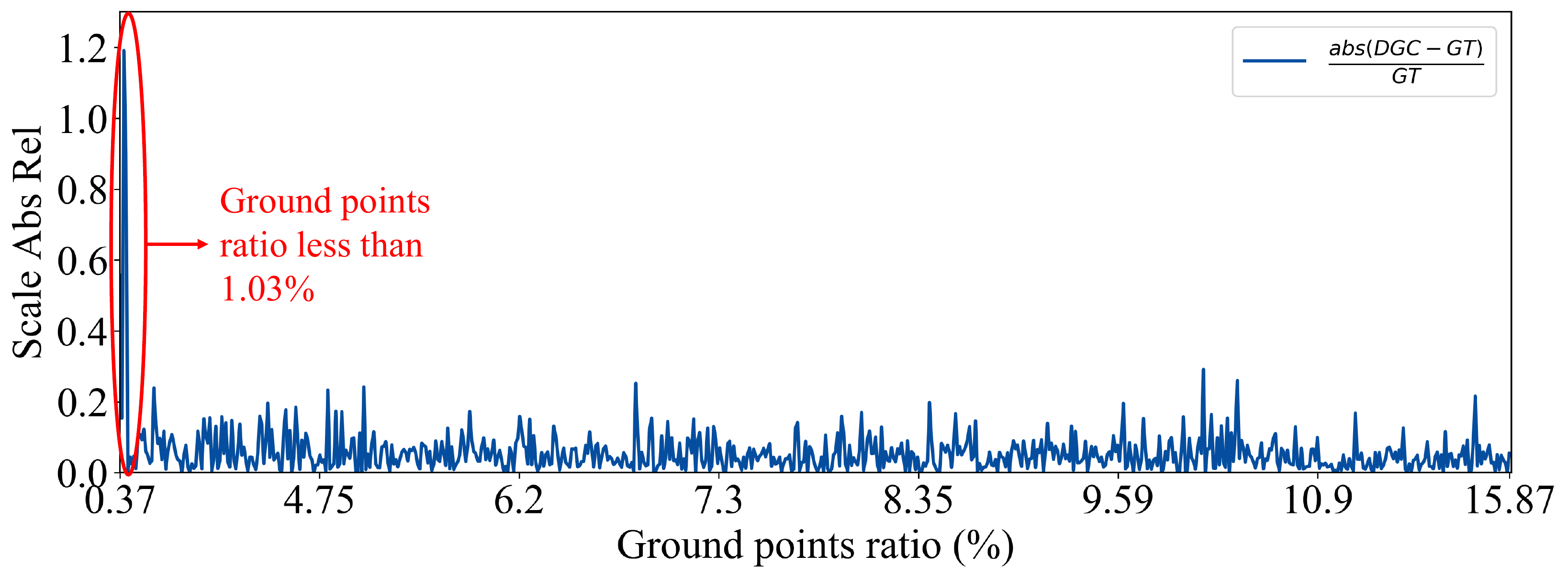}
	\caption{\textbf{Robustness evaluation of DGC scale recovery module} under different ground point ratios. The result shows that when detected ground points take up more than 1.03\% of all pixels, our proposed DGC module can perform comparable to GT scale recovery.}
	\label{fig:robustness-results}
\end{figure}

Since DGC scale recovery largely depends on the ground points extraction, the relationship of its performance and the proportion of ground points in a single frame should be carefully evaluated. We evaluate 697 test images in Eigen split and plot ground points ratio and corresponding scale error of each frame. Result is shown in Fig.~\ref{fig:robustness-results}, where the x-axis is ground point ratio and y-axis is $\frac{\text{DGC}-\text{GT}}{\text{GT}}$. It can be seen that when the ground point ratio is larger than 1.03\%, the proposed DGC module can perform uniformly and robustly comparable to GT scale recovery. But with extreme low ground points ratio, scale may be incorrectly estimated.

\begin{table}[h]
	\small
	\setlength{\tabcolsep}{8mm}
	\caption{\textbf{Speed performance of DNet.}}
	\label{tab:speed}
	\begin{center}
		\begin{tabular}{c c}
			\hline
			\textbf{Stage} & \textbf{Time consumption} \\
			\hline
			Inference & 50.0ms \\
			DGC scale recovery & 4.1ms\\
			\hline
		\end{tabular}
	\end{center}
\end{table}

\begin{figure*}[thpb]
	\centering
	\begin{minipage}{0.23\textwidth}
		\centering
		Input image\\
		\includegraphics[width=1\columnwidth]{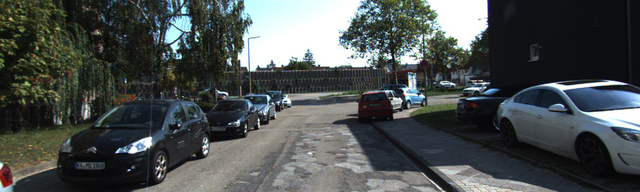}
		\\
		\includegraphics[width=1\columnwidth]{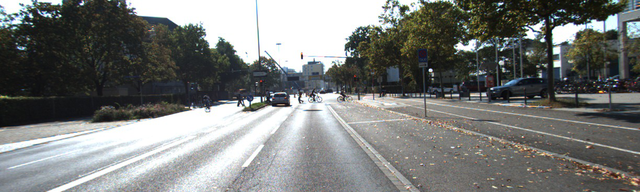}
		\\
		\includegraphics[width=1\columnwidth]{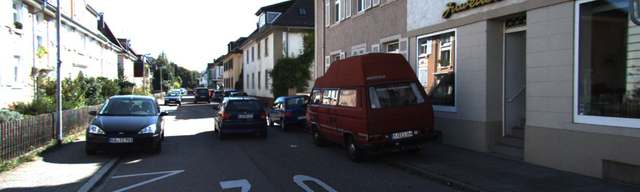}
	\end{minipage}
	\begin{minipage}{0.23\textwidth}
		\centering
		Depth estimation\\
		\includegraphics[width=1\columnwidth]{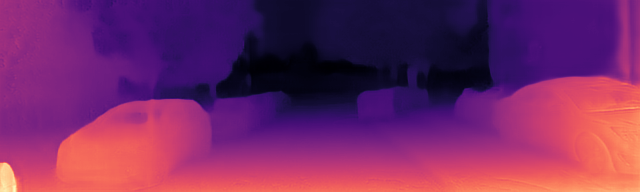}
		\\
		\includegraphics[width=1\columnwidth]{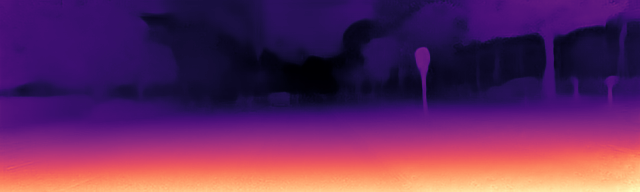}
		\\
		\includegraphics[width=1\columnwidth]{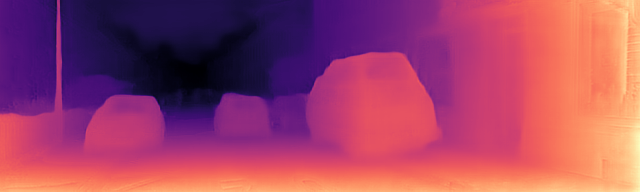}
	\end{minipage}	
	\begin{minipage}{0.23\textwidth}
		\centering
		Surface normal map\\
		\includegraphics[width=1\columnwidth]{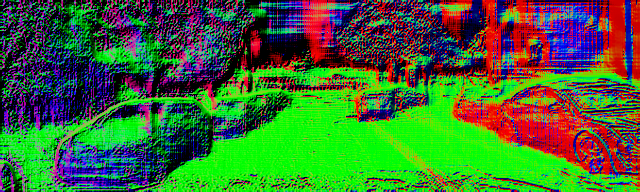}
		\\
		\includegraphics[width=1\columnwidth]{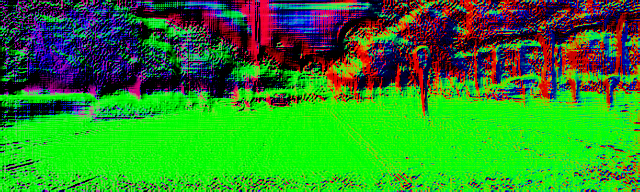}
		\\
		\includegraphics[width=1\columnwidth]{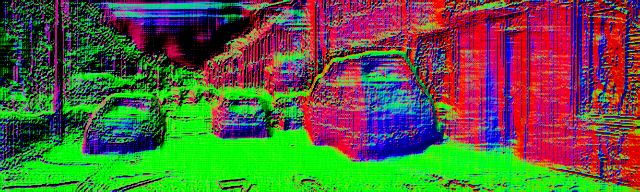}
	\end{minipage}
	\begin{minipage}{0.23\textwidth}
		\centering
		Ground point estimation\\
		\includegraphics[width=1\columnwidth]{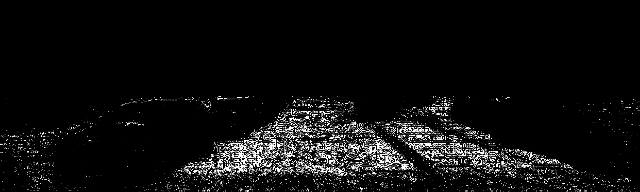}
		\\
		\includegraphics[width=1\columnwidth]{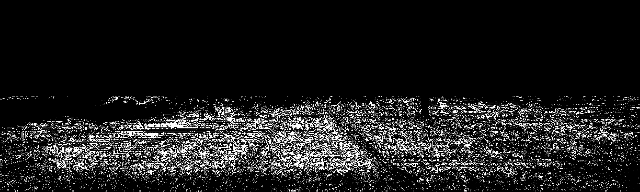}
		\\
		\includegraphics[width=1\columnwidth]{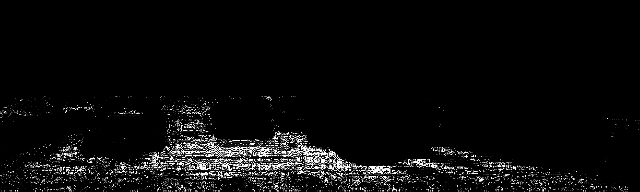}
	\end{minipage}
	\caption{\textbf{Qualitative results} of DNet absolute depth estimation result as well as components in DGC scale recovery module on KITTI 2015 Eigen Split. }
	\label{fig:qualitative_evaluation}
\end{figure*}

\begin{figure*}[thpb]
	\centering
	\includegraphics[width=1\linewidth]{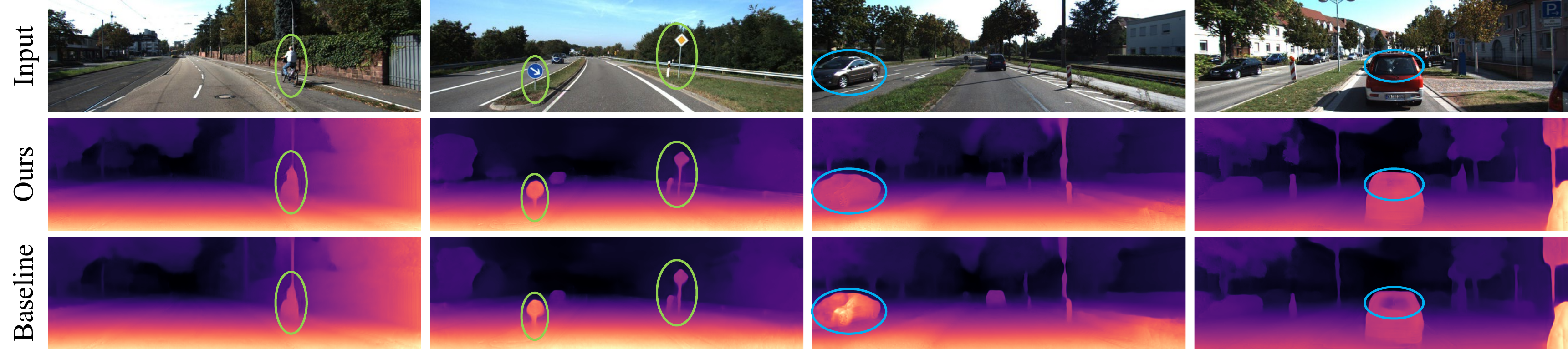}
	\caption{\textbf{Qualitative results} of our proposed DCP layer on KITTI 2015 Eigen Split. Compared to baseline, DNet with DCP layer is able to present more precise object boundary (green) and significantly reduce depth artifacts (blue).}
	\label{fig:DCP-layer-qualitative-results}
\end{figure*}

\subsection{Qualitative evaluation}

Qualitative results are demonstrated in Fig.~\ref{fig:qualitative_evaluation} and Fig.~\ref{fig:DCP-layer-qualitative-results}. Fig.~\ref{fig:qualitative_evaluation} shows the overall absolute depth estimation results as well as intermediate results such as surface normal and ground point mask. Fig.~\ref{fig:DCP-layer-qualitative-results} demonstrates intuitively the improvement brought by introducing DCP in comparison with baseline. It can be seen that object boundary is more precise and depth artifacts are to some extent eliminated. 

\subsection{Additional DGC and GT comprarisons}

\begin{figure}[!t]
	\centering
	\includegraphics[width=1\columnwidth]{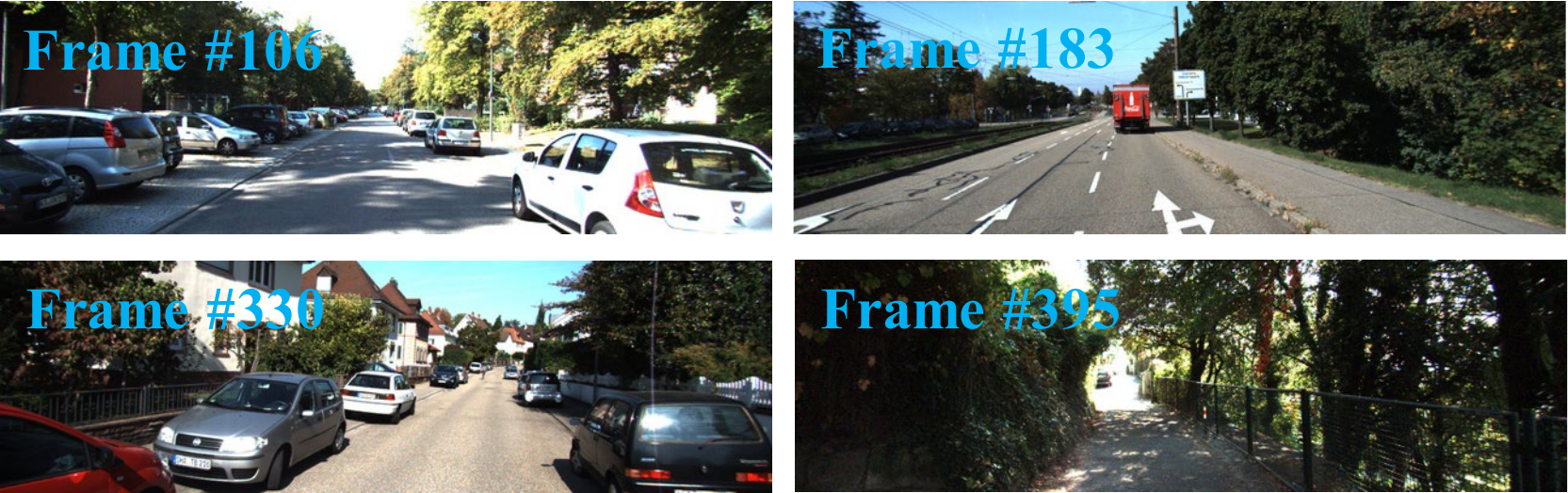}
	\caption{\textbf{Scenes where DGC scale recovery performs better than GT scale recovery.} It can be intuitively seen that ground point ratio are all relatively high in those scenes. }
	\label{fig:dgc-better}
\end{figure}

There are also results showing that in some cases, DGC scale recovery works even better than GT scale recovery, especially in those scenes, where ground point ratio is relatively large. Some example of those scenes can be seen in Fig.~\ref{fig:dgc-better}. The performance in those frames can be seen in Table~\ref{tab:dgc-better}. Surprisingly, in at least 31.7\% and at most 45.2\% of the frames, DGC scale recovery module performs better in terms of four metrics. Detailed result of the ratio of frames where DGC performs favorably against GT scale recovery can be seen in Table~\ref{tab:dgc-better-ratio}. 

\begin{table}[!t]
    \small
	\setlength{\tabcolsep}{1.5mm}
	\centering
	\caption{\textbf{Quantitative result in scenes listed in Fig.~\ref{fig:dgc-better}.} When ground point ratio is relatively high, DGC usually performs better than GT scale recovery in at least on metric.}
	\label{tab:dgc-better}%
	\begin{tabular}{|l|c|c|c|c|c|}
		\hline
		\multirow{2}*{Frame}& Scale & \multicolumn{4}{c|}{Lower is better} \\
		\cline{3-6}
		~ & Factor & Abs Rel & Sq Rel & RMSE & RMSE log \\
		\hline
		\#106 & GT & 0.195 & 1.443 & 6.416 & 0.320 \\
		\#106 & DGC & 0.110 & 1.105 & 5.799 & 0.319 \\
		\hline
		\#183 & GT & 0.194 & 1.175 & 5.888 & 0.231 \\
		\#183 & DGC & 0.127 & 0.995 & 5.745 & 0.229 \\
		\hline
		\#330 & GT & 0.211 & 1.100 & 4.138 & 0.270 \\
		\#330 & DGC & 0.144 & 0.844 & 4.174 & 0.271 \\
		\hline
		\#395 & GT & 0.353 & 2.181 & 5.837 & 0.418 \\
		\#395 & DGC & 0.273 & 1.754 & 5.834 & 0.470 \\
		\hline
	\end{tabular}%
\end{table}%

\begin{table}[!t]
    \small
	\setlength{\tabcolsep}{4mm}
	\centering
	\caption{\textbf{Ratio of the frames where DGC scale recovery performs better than GT in terms of different metrics.} It can be seen that especially in absolute relative errors, DGC performs better in many frames.}
	\label{tab:dgc-better-ratio}%
	\begin{tabular}{c c c c}
	    \hline
		\multicolumn{4}{c}{Evaluation metrics}\\
		\hline
		Abs Rel & Sq Rel & RMSE & RMSE log \\
		45.2\% & 38.5\% & 39.3\% & 31.7\%\\
		\hline
	\end{tabular}%
\end{table}%

\section{CONCLUSIONS}
In this work, a novel pipeline for self-supervised monocular absolute depth estimation is presented. DCP layer is proposed to generate hierarchical features for high resolution depth inferences, so that object boundary can be more accurate and depth artifacts can be better addressed. In order for the self-supervised monocular depth estimation to be more easily adapted to and used in autonomous driving applications, DGC module is introduced to perform absolute depth prediction without additional sensors and depth ground truth. Extensive experiments were conducted to demonstrate the effectiveness and robustness of the proposed DNet pipeline as well as DCP and DGC module. In future, this work provides intuition for better use of hierarchical features and can serve as the basis for further explorations of scale recovery methods.

\section*{ACKNOWLEDGMENT}
This research was supported in part by the National Natural Science Foundation of China under Grant U19A2069, the National Research Foundation Singapore through the Singapore MIT Alliance for Research and Technology's Future Urban Mobility IRG research programme, and the Singapore Land Transport Authority Urban Mobility Grand Challenge Funding Initiative.

{\small

 \bibliographystyle{IEEEtran}
}

\end{document}